\documentclass[letterpaper, 10 pt, conference]{ieeeconf}

\IEEEoverridecommandlockouts                              

\title{\LARGE \bf Lidar-Monocular Visual Odometry with Genetic Algorithm for Parameter Optimization}

\author{Adarsh Sehgal, Ashutosh Singandhupe, Hung Manh La, Alireza Tavakkoli, Sushil J. Louis
\thanks{Adarsh Sehgal, Ashutosh Singandhupe and Dr. Hung La are with the Advanced Robotics and Automation
(ARA) Laboratory. Dr. Alireza Tavakkoli is with the Computer Vision Laboratory (CVL). Dr. Sushil J. Louis is with the Evolutionary Computing Systems Lab, University of Nevada, Reno, NV 89557, USA. Corresponding author: Hung La, email: {\tt\small hla@unr.edu}}
\thanks{This material is based upon work supported by the National Aeronautics and Space Administration (NASA) Grant No. NNX15AI02H issued through the NVSGC-RI program under sub-award No. 19-21, and the RID program under sub-award No. 19-29, and the NVSGC-CD program under sub-award No. 18-54. This work is also partially supported by the Office of Naval Research under Grant N00014-17-1-2558.}
}

\author{Adarsh Sehgal, Ashutosh Singandhupe, Hung Manh La, Alireza Tavakkoli, Sushil J. Louis
\thanks{Adarsh Sehgal, Ashutosh Singandhupe and Dr. Hung La are with the Advanced Robotics and Automation
(ARA) Laboratory. Dr. Alireza Tavakkoli is with the Computer Vision Laboratory (CVL). Dr. Sushil J. Louis is with the Evolutionary Computing Systems Lab, University of Nevada, Reno, NV 89557, USA. Corresponding author: Hung La, email: {\tt\small hla@unr.edu}}
\thanks{This material is based upon work supported by the National Aeronautics and Space Administration (NASA) Grant No. NNX15AI02H issued through the NVSGC-RI program under sub-award No. 19-21, and the RID program under sub-award No. 19-29, and the NVSGC-CD program under sub-award No. 18-54. This work is also partially supported by the Office of Naval Research under Grant N00014-17-1-2558.}
}

\usepackage{amsmath}
\usepackage{algorithm}
\usepackage{algpseudocode}
\usepackage{graphicx}
\usepackage{subcaption}
\usepackage{lipsum}
\usepackage{multicol}
\usepackage{cite}
\graphicspath{{images/}}
\begin{document}

\maketitle
\thispagestyle{empty}
\pagestyle{empty}

\begin{abstract}
Lidar-Monocular Visual Odometry (LIMO), a odometry estimation algorithm, combines camera and LIght Detection And Ranging sensor (LIDAR) for visual localization by tracking camera features as well as features from LIDAR measurements, and it estimates the  motion using Bundle Adjustment based on robust key frames. For rejecting the outliers,  LIMO uses semantic labelling and weights of the vegetation landmarks. A drawback of LIMO as well as many other odometry estimation algorithms is that it has many parameters that need to be manually adjusted according to the dynamic changes in the environment in order to decrease the translational errors. In this paper, we present and argue the use of Genetic Algorithm to optimize parameters with reference to LIMO and maximize LIMO's localization and motion estimation performance. We evaluate our approach on the well known KITTI odometry dataset and show that the genetic algorithm helps LIMO to reduce translation error in different datasets.

\end{abstract}

\section{INTRODUCTION and RELATED WORK}
Motion estimation has long been a popular subject of research in which many techniques have been developed over the years \cite{cremers2017direct}. Much work has been done related to Visual Simultaneous Localization and Mapping (VSLAM), also referred to as Visual Odometry \cite{taketomi2017visual}, which simultaneously estimates the motion of the camera and the 3D structure of the observed environment. A recent review of SLAM techniques for autonomous car driving can be found in \cite{slam2019review}. Bundle
Adjustment is the most popular method for VSLAM. Bundle Adjustment is a procedure of minimizing the re-projection error between the observed point (landmarks in reference to LIMO) and the predicted points. Recent developments make use of offline VSLAM for mapping and localization \cite{hartley2003multiple, szeliski2010computer, sons2015multi}. 

\begin{figure}[h!]
\centering
\hspace*{-1cm}
  \begin{subfigure}[b]{0.8\linewidth}
    \includegraphics[width=\linewidth,height=4cm]{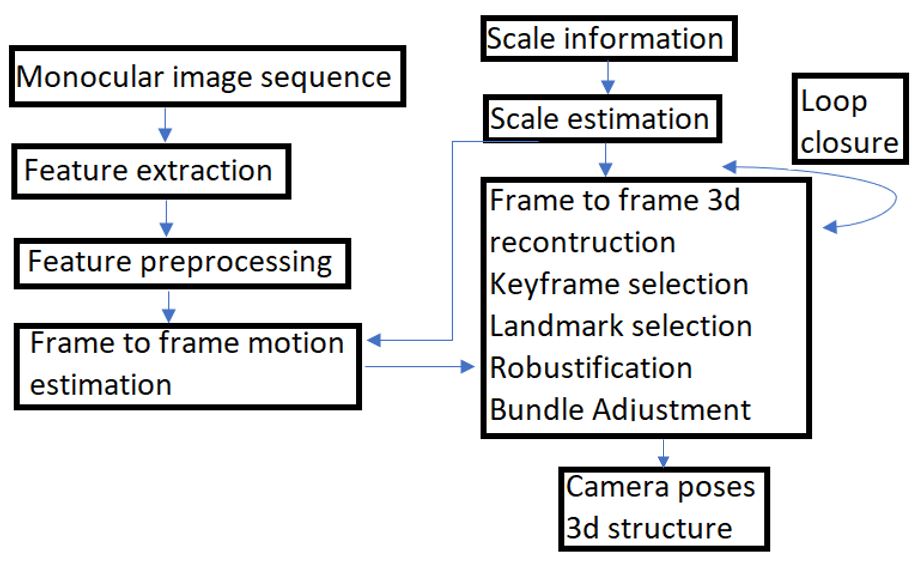}
  \end{subfigure}
  \caption{VSLAM pipeline. The input is temporal sequence of images, and the system outputs a sparse reconstruction of the observed environment and the camera poses.  \cite{mur2017orb} \cite{grater2015robust} \cite{nutzi2011fusion}  \cite{graeter2018limo}. In this work, LIMO does not perform loop closure.}
  \label{fig:VSLAMpipeline}
\end{figure}

Figure \ref{fig:VSLAMpipeline} illustrates the structure of the VSLAM pipeline \cite{graeter2018limo}. 
Algorithms such as  ROCC \cite{buczko2016flow} and SOFT \cite{cvivsic2015stereo} rely on pre-processing and feature extraction which is in very contrast to most of the methods that obtain scale information from a camera placed at a different viewpoint\cite{sons2015multi, mur2017orb, geiger2011stereoscan, cvivsic2017soft}.
The former mentioned algorithms (SOFT and ROCC) extract robust and precise features and select them using special techniques, and has managed to attain high performance on the KITTI Benchmark \cite{geiger2013vision}, even without Bundle Adjustment.

The major disadvantage of stereo camera is it's reliance on extrinsic camera calibration. It was later observed that the performance can be enhanced by learning a compensation of the calibration bias through a deformation field \cite{krevso2015improving}. 
LIght Detection And Ranging sensor LIDAR-camera calibration is also an expanding topic of research \cite{geiger2012automatic, grater2016photometric} with accuracy reaching a few pixels.
Previous work has been done with VSLAM and LIDAR \cite{xu2018combining, zhang2015visual, caselitz2016monocular, zhang2014loam}. Lidar-Monocular Visual Odometry (LIMO) \cite{graeter2018limo}, uses feature tracking capability of the camera and combines it with depth measurements from a LIDAR but suffers from translation and rotation errors. Later on, we describe our approach for increasing LIMO's robustness to translation errors. 

LIMO takes advantage of the depth information from LIDAR which is to be used for feature detection in the image. Outliers are rejected if do not meet the local plane assumptions,  and points on the ground plane are treated for robustness. As illustrated in figure \ref{fig:VSLAMpipeline}, in the VSLAM pipieline, the depth information is fused with monocular feature detection techniques. Another approach is taken for prior estimation, landmark selection and key frame selection to fulfill real time constraints. Unlike the approach in \cite{zhang2015visual}, LIMO does not use any LIDAR-SLAM algorithms such as Iterative Closest Point (ICP). The major drawback of LIMO is that it has many parameters, which needs to be manually tuned. LIMO suffers from translation and rotation errors even more than existing algorithms such as Lidar Odometry and Mapping (LOAM) \cite{zhang2014loam} and Vision-Lidar Odometry and Mapping (V-LOAM) \cite{balazadegan2016visual}. 
Typically, researchers tune parameters (in LIMO as well) in order to minimize these errors but there always exists the possibility of finding better parameter sets that may be optimized for specific camera and LIDAR hardware or for specific scenarios.  
Hence, there is a need to use optimization algorithms to increase LIMO's performance. In this paper, we propose using a genetic algorithm (GA) to efficiently search the space of possible LIMO parameter values to find precise parameter that maximizes performance. Our experiments with this new GA-LIMO algorithm show that GA-LIMO performs little better than stock LIMO.

Much empirical evidence shows that evolutionary computing techniques such as Genetic Algorithms (GAs) work well as function optimizers in poorly-understood, non-linear, discontinuous spaces \cite{holland1992adaptation, goldberg1988genetic, mitchell1998introduction, gibb2018genetic, tavakkoli2007genetic}. GAs \cite{davis1991handbook, kalyanmoy2002fast} and the GA operators of crossover and mutation \cite{poon1995genetic} have been tested on numerous problems. Closer to our research, GAs have been applied to early SLAM optimization problems \cite{duckett2003genetic}, mobile localization using ultrasonic sensors \cite{moreno2002genetic} \cite{La_SMC2009}, and in deep reinforcement learning \cite{deep2019geneticalgo}. This provides good evidence for GA efficacy on localization problems, and our main contribution in this paper is a demonstration of smaller translation error when using a GA to tune LIMO parameter values compared to the stock LIMO algorithm \cite{graeter2018limo}. Our experiments show that translation error is non-linearly related to LIMO parameters, that is, translation error can vary non-linearly based on the values of the LIMO's parameters. The following sections describe the LIMO, the GA and GA-LIMO algorithms. We then show results from running LIMO with GA tuned parameters on the KITTI odometry data sequences \cite{Geiger2012CVPR}.

This paper is organized as follows: Section 2 describes the LIMO and GA algorithms. Section 3 describes the combined GA-LIMO algorithm. Section 4 then specifies our experiments and provides results. The last section delivers conclusions and possible future work.

\section{BACKGROUND}

In this section, we present prior work related to our GA-LIMO algorithm. We first describe the VSLAM pipeline and then the LIMO algorithm.

\subsection{Feature extraction and pre-processing}

Figure \ref{fig:VSLAMpipeline} shows feature extraction's procedure in the pipeline. Feature extraction consists of tracking the features and associating the features using the Viso2 library. \cite{geiger2011stereoscan} It is further used to implement feature tracking which comprises of non-maximum suppression, sub-pixel refinement and outlier rejection by flow. Deep learning is used to reject landmarks that are moving objects. The neighborhood of the feature point in a semantic image \cite{cordts2016cityscapes} is scanned, and if the majority of neighboring pixels categorize to a dynamic class, like vehicle or pedestrian, the landmark is excluded.

\subsection{Scale Estimation}

For scale estimation, the detected feature points from camera is mapped to the depth extracted from LIDAR. LIMO uses a one shot depth estimation approach. Initially LIDAR point cloud is transformed into the camera frame and then it is projected onto the image plane. In detail, the following steps are executed for every feature point $f$:

\begin{enumerate}
   \item A region of interest is selected around $f$ which is a set $F$ consisting of projected LIDAR points.
   \item A new set called foreground set $F_{seg}$ is created by segmenting the elements of $F$.
   \item The elements of $F_{seg}$ are fitted with a plane $p$. A special fitting algorithm is used in case $f$ belongs to the ground plane.
   \item To estimate the depth, $p$ is intersected with the line of sight corresponding to $f$ .
   \item For the previous estimated depth a test is performed. Depth estimates that are more than 30m are rejected since they could be uncertain. In addition, the angle between the line of sight of the feature point and the normal of the plane must be smaller than a threshold. 
\end{enumerate}

From the point clouds, neighborhoods for ordered point clouds can be selected directly. However, projections of the LIDAR points on the image are used, and the points within a rectangle in the image plane around $f$ are selected in case the point clouds are unordered. Before the plane estimation is performed, the foreground $F_{seg}$ is segmented. In the next step, a histogram of depth having a fixed bin width of $h = 0.3$m is created and interpolated with elements in $F$. LIDAR points of the nearest bin is used to perform segmentation using all detected feature points. For estimating the local surface around $f$ precisely, fitting the plane to $F_{seg}$ can help. Three points are chosen from the points in $F_{seg}$, which traverse the triangle $F_{\Delta}$ with maximum area. Depth estimation is avoided if the area of $F_{\Delta}$ is too small, to evade incorrectly estimated depth. 

However, the above technique cannot be used to estimate the the depth of points on the ground plane because LIDAR has a lower resolution in the perpendicular direction than in a level direction. A different approach is followed to enable depth estimation for a relevant ground plane. For solving this, RANSAC with refinement is used on the LIDAR point cloud to extract the ground plane\cite{szeliski2010computer}. In order to estimate feature points on the road, points that corresponds to the ground plane are segmented. Outliers are extracted by allowing only local planes that lie close to the ground plane.   

\subsection{Frame to Frame Odometry}

Perspective-n-Point-Problem \cite{szeliski2010computer} serves as the starting point of the frame to frame motion estimation. 
\begin{gather} 
\underset{x,y,z,\alpha,\beta,\gamma}{argmin}  \underset{i}\sum  \|\varphi_{i,3d\rightarrow 2d}\|_{2}^{2}     
\label{first}
\end{gather}
\begin{gather} 
\varphi_{3d\rightarrow2d} = \bar{p}_{i} - \pi(p_{i}, P(x,y,z,\alpha,\beta,\gamma)),
    \label{second}
\end{gather}
where $\bar{p_{i}}$ is the observed feature point in the current frame, $p_{i}$ is the 3D-point corresponding to $\bar{p_{i}}$, the transform from the previous to the current frame is denoted by freedom $P(x,y,z,\alpha,\beta,\gamma)$, which has three translation and three rotation degrees of freedom. While $\pi(...)$ is the projection function from the 3D to 2D domain. The extracted features with valid estimated depth depth from the environments that has low structure and large optical flow may be very small. LIMO introduces epipolar error as $\varphi_{2d\rightarrow2d}$ \cite{hartley2003multiple}.
\begin{gather} 
     \varphi_{2d\rightarrow2d} = \bar{p}_iF(\frac{x}{z},\frac{y}{z},\alpha,\beta,\gamma)\bar{p}_i,
    \label{third}
\end{gather}
where fundamental matrix $F$ can be calculated from the intrinsic calibration of the camera and from the frame to frame motion of the camera. LIMO suggests the loss function to be Cauchy function \cite{hartley2003multiple}:  $\rho_s(x)=a(s)^2.log(1+\frac{x}{a(s)^2})$, where $a(s)$ is the fix outlier threshold. For frame to frame motion estimation, the optimization problem can be denoted as:
\begin{gather} 
    \underset{x,y,z,\alpha,\beta,\gamma}{argmin}  \underset{i}\sum  \rho_{3d\rightarrow2d}(\|\varphi_{i,3d\rightarrow 2d}\|_{2}^{2})+\rho_{2d\rightarrow2d}(\|\varphi_{i,2d\rightarrow2d}\|_2^2).
    \label{forth}
\end{gather}

\subsection{Backend}

LIMO proposes a Bundle Adjustment framework based on keyframes , with key components as selection of keyframes, landmark selection, cost functions and robustification measures. The advantage with this approach is that it retains the set that carries information which are required for accurate pose estimation as well as excludes the unnecessary measurements. Keyframes are classified as required, rejected and sparsified keyframes. Required frames are crucial measurements. Frame rejection is done when the vehicle does not move. The remaining frames are collected, and the technique selects frames every 0.3s. Finally in keyframe selection, length of optimization window is chosen. 

An optimal set of landmarks should be well observable, small, free of outliers and evenly distributed. Landmark selection divides all landmarks into  three bins, near, middle and far, each of which have fixed number of landmarks selected for the Bundle Adjustment. Weights of the landmarks are then determined based on the semantic information. The estimated landmark depth is taken into consideration by an additional cost function, \begin{gather} 
    \xi_{i,j}(i_i,P_j) = 
    \begin{cases}
        0, \; \; if\;l_i\;has\; no\; depth\;estimate \\
        \hat{d}_{i,j} - \begin{bmatrix} 0 & 0 & 1 \end{bmatrix} \tau(l_i,P_j),\; \; else,
    \end{cases}
    \label{fifth}
\end{gather}
where $l_i$ denotes the landmark, $\tau$ mapping from world frame to camera frame and $\hat{d}$ denotes the depth estimate. The indices $i,j$ denote combination of landmark-pose. A cost function $\nu$ punishes deviations from the length of translation vector,
\begin{gather} 
    \nu(P_1,P_0) = \hat{s}(P_1,P_0)-s,
    \label{sixth}
\end{gather}
where $P_0$, $P_1$ are the last two poses in the optimization window and $\hat{s}(P_0,P_1) = \|translation(P^{-1}P_1)\|_2^2$, where $s$ is a constant with value of $\hat{s}(P_1,P_0)$ before optimization.

While outliers need to be removed because they do not let Least-Square methods to converge \cite{torr2000mlesac, torr2004invariant}, semantics and cheirality only does preliminary outlier rejection. The LIMO optimization problem can now be formulated as:
\begin{gather} 
    \underset{P_j\in P,l_i\in L,d_i\in D}{argmin} w_0\|\nu(P_1,P_0\|_2^2) + \nonumber\\
    \underset{i}{\sum}\underset{j}{\sum} w_1\rho_{\phi}(\|\phi_{i,j}(l_i,P_i)\|_2^2) + w_2\rho_{\xi}(\|\xi_{i,j}(l_i,P_j)\|_2^2),
    \label{seventh}
\end{gather}
where $\phi_{i,j}(l_i,P_j) = \bar{l}_{i,j} - \pi(l_i,P_j)$ is the re-projection error, and weights $w_0$, $w_1$ and $w_2$ are used to scale the cost functions to the same order of magnitude.

\subsection{Genetic Algorithm (GA)}

\textit{GA} \cite{holland1992adaptation, davis1991handbook, polyak1992acceleration, goldberg1988genetic} were designed to  search poorly-understood spaces, where exhaustive search may not be feasible (because of search space and time space complexity), and where other search techniques perform poorly. A GA as a function optimizer tries to maximize a fitness tied to the optimization objective. In general, evolutionary computing algorithms, and specifically GAs have had verifiable success on a diversity of difficulty, non-linear,  design and optimization problems. GAs usually start with a  randomly initialization population of candidate solution encoded in a string. Selection operators then focus search on higher fitness areas of search space whereas crossover and mutation operators generate new candidate solutions. We next explain the specific GA used in this paper.

\section{GA-LIMO algorithm}

In this section, we present the main contribution of our paper. The specific GA searches through the space of parameter values used in LIMO for the values that maximizes the performance and minimizes the translation error as a result of pose estimation. We are targeting the following parameters: outlier rejection quantile $\delta$; maximum number of landmarks for near bin $\epsilon_{near}$; maximum number of landmarks for middle bin $\epsilon_{middle}$; maximum number of landmarks for far bin $\epsilon_{far}$ and weight for the vegetation landmarks $\mu$. As described in the background section, rejecting outliers, $\delta$, plays an important role in converging to minimum, the weight of outlier rejection thus has notable impact on the translation error. The landmarks are categorized into three bins, which also have great significance in translation error calculation. Trees that have a rich structure results in feature points that are good to track, but they can move. So, finding an optimal weight for vegetation can significantly reduce translation error. $\delta$ and $\mu$ range from 0 to 1, while $\epsilon_{near}$, $\epsilon_{middle}$ and $\epsilon_{far}$ range from 0 to 999. We have set these ranges based on early experimental results. 

\begin{algorithm}
\caption{GA-LIMO}\label{euclid}
\begin{algorithmic}[1]
\State Choose population of $n$ chromosomes
\State Set the values of parameters into the chromosome
\State Run LIMO with the GA selected parameter values
\For{all chromosome values} 
    \State Run LIMO on KITTI odometry data set sequence 01
    \State Compare LIMO estimated poses with ground truth
    \State Translation error $\sigma_1$ is found
    \State Run LIMO on KITTI odometry data set sequence 04
    \State Compare LIMO estimated poses with ground truth
    \State Translation error $\sigma_4$ is found
    \State Average error $\sigma_{avg} = \frac{\sigma_1+\sigma_4}{2}$
    \State \textbf{return} $1/\sigma_{avg}$
\EndFor
\State Perform Uniform Crossover
\State Perform Flip Mutation at rate 0.1
\State Repeat for required number of generations for optimal solution
\end{algorithmic}
\end{algorithm}

Our experiments show that adjusting the values of parameters did not decrease or increase the translation error in a linear or easily appreciable pattern. So, a simple hill climber will probably not do well in finding optimized parameters. We thus use a GA to optimize these parameters.

Algorithm \ref{euclid} explains the combination of LIMO with the GA, which uses a population size of 50 runs for 50 generations. We used ranking selection \cite{goldberg1991comparative} to select the parents for crossover and mutation. Rank selection probabilistically selects higher ranked (higher fitness) individuals. Unlike fitness proportional selection, ranking selection pays attention to the existence of a fitness difference rather than also to the magnitude of fitness difference. Children are generated using uniform crossover \cite{syswerda1989uniform}, which are then mutated using flip mutation \cite{goldberg1988genetic}. Chromosomes are binary encoded with concatenated parameters. $\delta$ and $\mu$ are considered up to three decimal places, which means a step size of 0.001, because changes in values of parameters cause considerable change in translation error. All the parameters require 11 bits to represent their range of values, so we have a chromosome length of $55$ bits, with parameters arranged in the order: $\delta$, $\epsilon_{near}$, $\epsilon_{middle}$, $\epsilon_{far}$, $\mu$.

The algorithm starts with randomly generating a population of $n$ individuals. Each chromosome is sent to LIMO to evaluate. LIMO evaluates the parameter set represented by this individual by using those parameters to run on the KITTI dataset \cite{geiger2013vision}. The KITTI benchmarks are well known and provide the most popular benchmark for Visual Odometry and VSLAM. This dataset has rural, urban scenes along with highway sequences and provides gray scale images, color images, LIDAR point clouds and their calibration. Most LIMO configurations are as in \cite{graeter2018limo}. In our work, we focus on two sequences in particular: sequence 01 and 04. Sequence 01 is a highway scenario, which is challenging because only a road can be used for depth estimation. Sequence 04 is an urban scenario, which has a large number of landmarks for depth estimation. We consider both sequences for each GA evaluation because we want a common set of parameters that work well with multiple scenes.   

The fitness of each chromosome is defined as the inverse of translation error. This translates the minimization of translation error into a maximization of fitness as required for GA optimization. Since each fitness evaluation takes significant time, an exhaustive search of the $2^{55}$ size search space in not possible, hence our using the GA. During a fitness evaluation, the GA first runs the LIMO with sequence 01. It then compares the LIMO estimated poses with ground truth (also found in \cite{geiger2013vision}) and finds the translation error using the official KITTI metric \cite{geiger2013vision}. The same steps are followed for sequence 04. The fitness value of each chromosome is the average of the inverse translation errors from the two sequences.
\begin{gather} 
    \sigma_{avg} = \frac{\sigma_1+\sigma_4}{2}
    \label{eighth}.
\end{gather}
Selected chromosomes (ranked selection) are then crossed over and mutated to create new chromosomes to form the next population. This starts the next GA iteration of evaluation, selection, crossover, and mutation. The whole system takes significant time since we are running $50 * 50 = 2500$ LIMO evaluations to determine the best parameters. The next section shows our experiments with individual and combined sequences, with and without the GA. Our results show that the GA-LIMO performs better than the results of LIMO \cite{graeter2018limo}.

\section{EXPERIMENT and RESULTS}

In this section we show our experiments with individual KITTI sequences, a combination of sequences, and overall results. First, we run the GA-LIMO with sequences 01 and 04 separately. We show the translation error and the error mapped onto trajectory, compared to the ground truth (reference) \cite{Geiger2012CVPR}. We then, show our results when GA-LIMO runs with evaluations on both sequences 01 and 04. Finally, we compare the values of parameters found by GA-LIMO versus LIMO.

\begin{figure*}
\centering
  \begin{subfigure}[b]{\linewidth}
  \includegraphics[width=\textwidth,height=4cm]{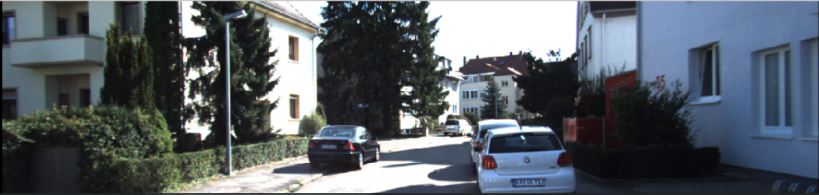}
  \end{subfigure}
  \caption{Camera data while GA-LIMO is in action.}
  \label{fig:cameradata}
\end{figure*}

\begin{figure*}
\centering
  \begin{subfigure}[b]{\linewidth}
  \includegraphics[width=0.245\linewidth,height=4cm]{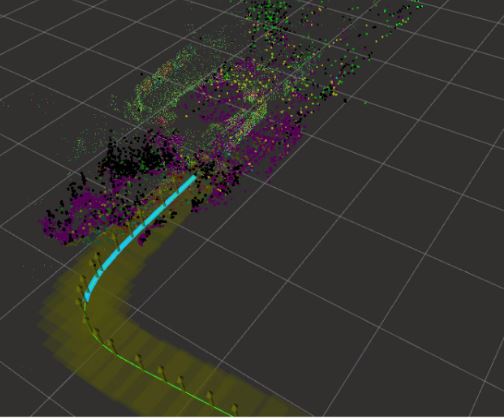}
  \includegraphics[width=0.245\linewidth,height=4cm]{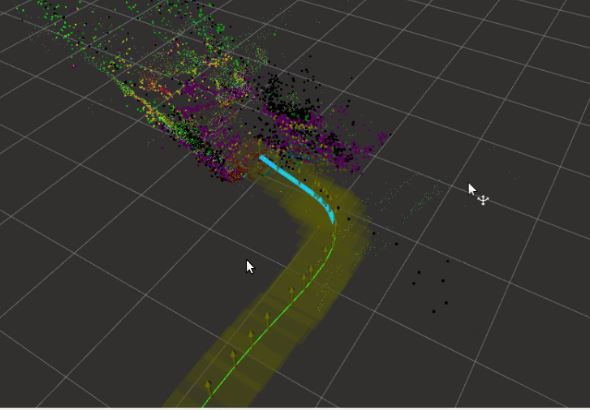}
  \includegraphics[width=0.245\linewidth,height=4cm]{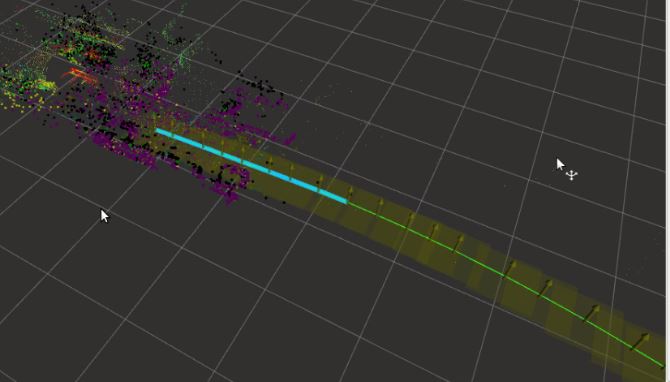}
  \includegraphics[width=0.245\linewidth,height=4cm]{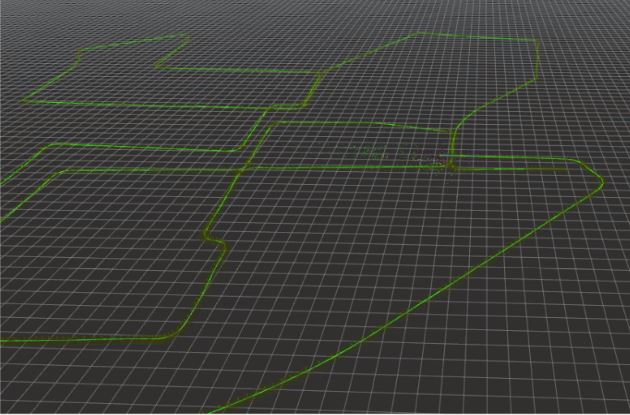}
  \end{subfigure}
  \caption{GA-LIMO estimating the pose.}
  \label{fig:lidardata}
\end{figure*}

\begin{figure}[h!]
\centering
\hspace*{-1.5cm}
  \begin{subfigure}[b]{0.8\linewidth}
   \centering
    \includegraphics[width=\linewidth,height=6.5cm]{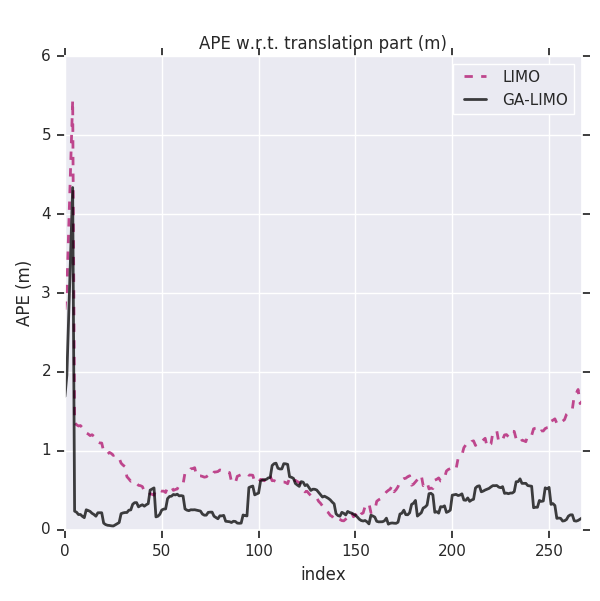}
    \caption{Translation error comparison over the poses.}
    \label{fig:rmse_compare_seq_04_individual}
  \end{subfigure}
  \begin{subfigure}[b]{\linewidth}
  \includegraphics[width=7.7cm,height=6.5cm]{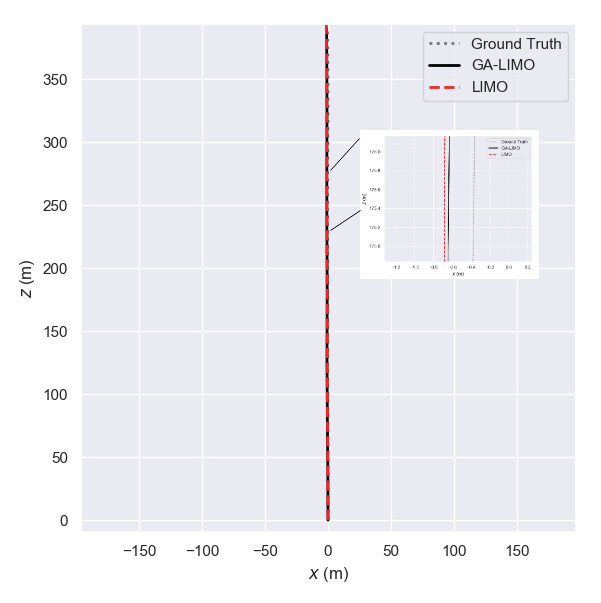}
  \caption{Trajectory comparison for sequence 04, when GA-LIMO was run on this sequence individually (algorithm \ref{individual}).}
  \label{fig:individual_seq_04_maps}
  \end{subfigure}
  \caption{Results comparison for sequence 04 (algorithm \ref{individual}). LIMO has 1.01\% translation error, while GA-LIMO has about half this error with 0.56\%.}
  \label{fig:sequence_04_individual}
\end{figure}

\begin{algorithm}
\caption{GA-LIMO individual}\label{individual}
\begin{algorithmic}[1]
\State Choose population of $n$ chromosomes
\State Set the values of parameters into the chromosome
\State Run LIMO with the GA selected parameter values
\For{all chromosome values} 
    \State Run LIMO on individual KITTI odometry dataset sequence
    \State Compare LIMO estimated poses with ground truth
    \State Translation error $\sigma$ is found
    \State \textbf{return} $1/\sigma$
\EndFor
\State Perform Uniform Crossover
\State Perform Flip Mutation at rate 0.1
\State Repeat for required number of generations to find optimal solution
\end{algorithmic}
\end{algorithm}

\begin{figure}[htbp]
\centering
\hspace*{-1cm}
  \begin{subfigure}[b]{0.8\linewidth}
   \centering
    \includegraphics[width=\linewidth,height=6.5cm]{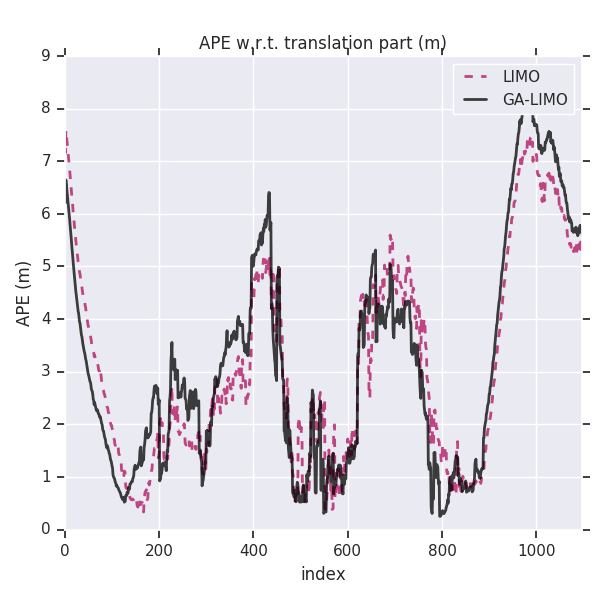}
    \caption{Translation error comparison over the poses.}\label{fig:rmse_compare_seq_01}
  \end{subfigure}
\begin{subfigure}[b]{\linewidth}
  \includegraphics[width=7.7cm,height=6.5cm]{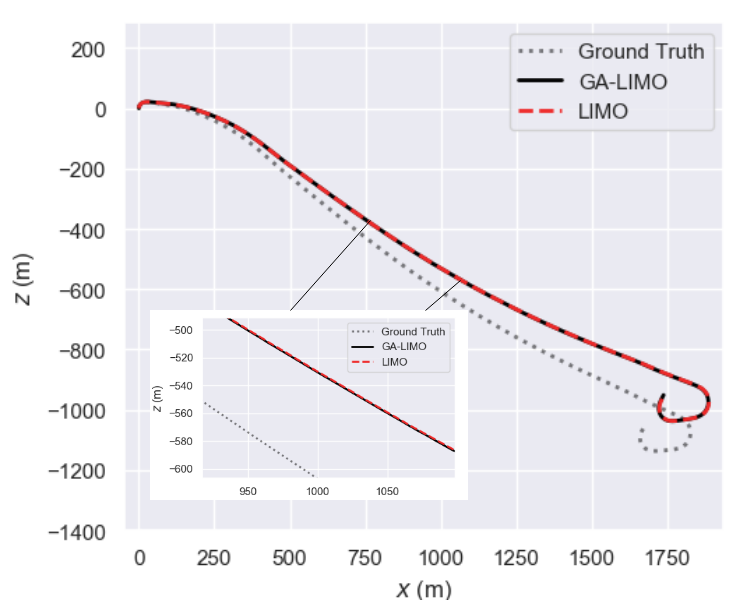}
  \caption{Trajectory comparison.}
  \label{fig:individual_seq_01_maps}
    \end{subfigure}
  \caption{Results comparison for sequence 01 (algorithm \ref{individual}). LIMO has 3.71\% translation error, while GA-LIMO has 3.8\%.}
  \label{fig:sequence_01_individual}
\end{figure}

Figure \ref{fig:cameradata} shows camera data while GA-LIMO is estimating the pose from that data in figure \ref{fig:lidardata}. Figure \ref{fig:sequence_04_individual} compares LIMO performance with GA-LIMO on sequence 04. Here the GA was run on this sequence individually to find the optimal parameters, as in algorithm \ref{individual}. Absolute Pose Error (APE) and Root Mean Squared Error (RMSE) are one of the important measures \cite{sturm2012benchmark}. The translation error for each sequence is the RMSE calculated with respect to ground truth. Figure \ref{fig:rmse_compare_seq_04_individual} compares the translation error over the poses, while figure \ref{fig:individual_seq_04_maps} compares the error mapped onto the trajectory with the zoomed in trajectory. Table \ref{table:1} compares the values of parameters for LIMO and GA-LIMO. Our results show that the GA-LIMO trajectory is closer to ground truth compared to LIMO. We found that the translation error was 0.56\% with GA-LIMO, in contrast to 1.01\% with LIMO.

\begin{table}[h!]
\centering
\begin{tabular}{||c c c||} 
 \hline
 Parameters & LIMO & GA-LIMO \\ [0.5ex] 
 \hline\hline
 $\delta$ & 0.95 & 0.986\\ 
 $\epsilon_{near}$ & 400 & 999\\
 $\epsilon_{middle}$ & 400 &  960\\
 $\epsilon_{far}$ & 400 & 859\\ 
 $\mu$ & 0.9 &  0.128\\[1ex] 
 \hline
\end{tabular}
\caption{LIMO vs GA-LIMO values of parameters when GA was run on LIMO with sequence 04 individually.}
\label{table:1}
\end{table}

\begin{figure}[htbp]
\centering
  \begin{subfigure}[b]{\linewidth}
  \centering
    \includegraphics[width=\linewidth,height=6.5cm]{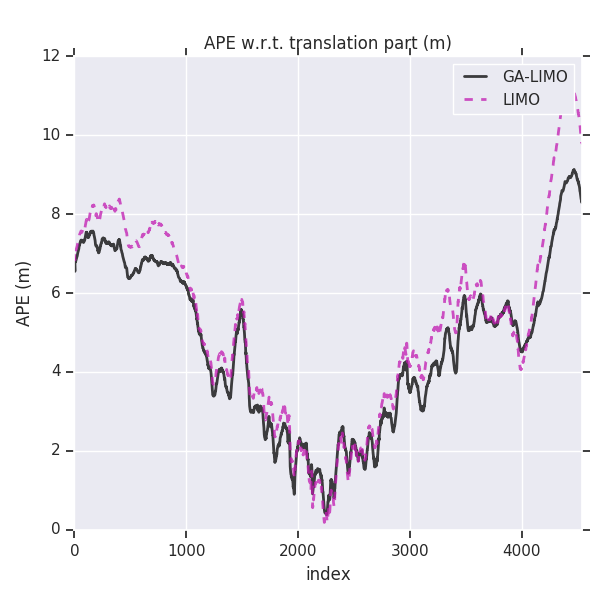}
    \caption{Translation error comparison over the poses for sequence 00. LIMO has 5.77\% translation error while GA-LIMO has 5.13\%.}
    \label{fig:map_01_04}
  \end{subfigure}
  \begin{subfigure}[b]{\linewidth}
   \centering
    \includegraphics[width=\linewidth,height=6.5cm]{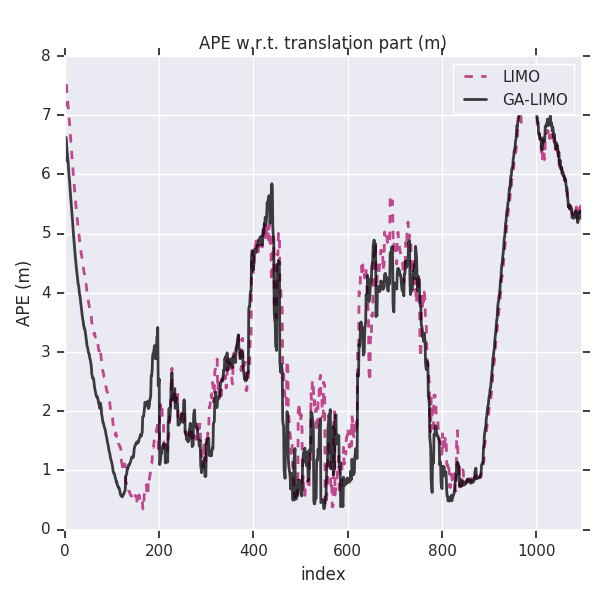}
    \caption{Translation error comparison over the poses for sequence 01. LIMO has 3.71\% translation error while GA-LIMO has 3.59\%.}
    \label{fig:combined_seq_01_compare}
  \end{subfigure}
  \begin{subfigure}[b]{\linewidth}
  \centering
    \includegraphics[width=\linewidth,height=6.5cm]{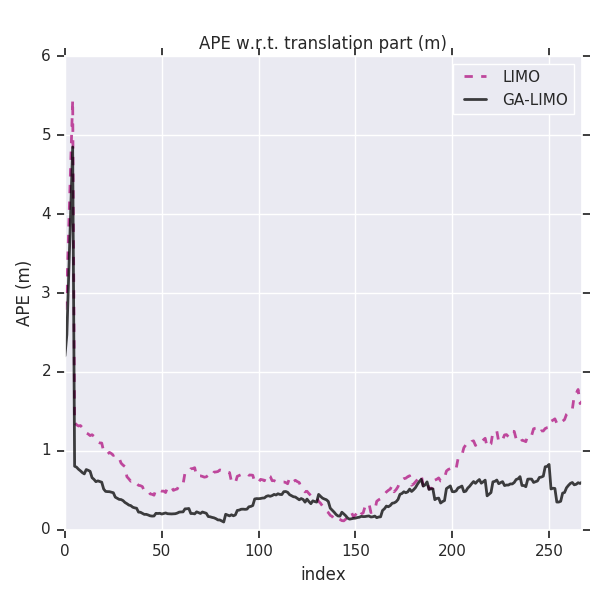}
    \caption{Translation error comparison over the poses for sequence 04. LIMO has 1.01\% translation error while GA-LIMO has 0.65\%.}
    \label{fig:combined_seq_04_compare}
  \end{subfigure}
  \caption{The parameters are found using GA-LIMO using combination of sequences 01 and 04 (Algorithm \ref{euclid}). These parameters are then tested on three sequences. In all three sequences GA-LIMO performs better than LIMO.}
  \label{fig:sequence_04_01}
\end{figure}

Figure \ref{fig:sequence_01_individual} compares the performance of LIMO with GA-LIMO when the system is run on just sequence 01. Here first, the GA was run on sequence 01 (Algorithm \ref{individual}) and the optimal parameters were used to test the same sequence. Table \ref{table:2} compares the original and GA found parameter values. Figure \ref{fig:rmse_compare_seq_01} compares translation error, while figure \ref{fig:individual_seq_01_maps} shows the error mapped onto the trajectory for LIMO and GA-LIMO. As shown in the zoomed in figure \ref{fig:individual_seq_01_maps}, GA-LIMO is closer to the ground truth. The translation error for LIMO is found to be around 3.71\% and 3.8\% in case of GA-LIMO, with sequence 01. GA found parameters that did not out perform the original parameters, when GA-LIMO was run on just sequence 01. 

\begin{figure}[htbp]
\centering
  \begin{subfigure}[b]{\linewidth}
  \includegraphics[width=\linewidth,height=6.2cm]{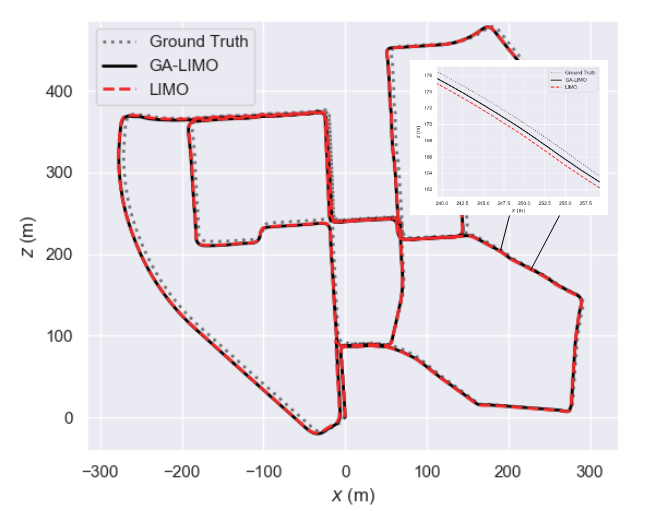}
  \label{fig:map_compare_seq_00}
  \caption{Sequence 00 trajectories showing GA-LIMO closer to ground truth.}
  \end{subfigure}
  \begin{subfigure}[b]{\linewidth}
  \includegraphics[width=\linewidth,height=6.2cm]{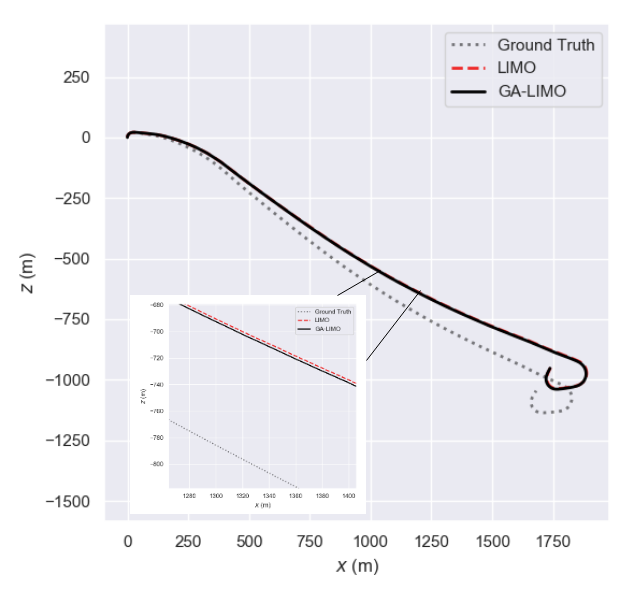}
  \label{fig:map_compare_seq_01}
  \caption{Sequence 01 trajectories showing GA-LIMO closer to ground truth.}
  \end{subfigure}
  \begin{subfigure}[b]{\linewidth}
  \includegraphics[width=\linewidth,height=6.2cm]{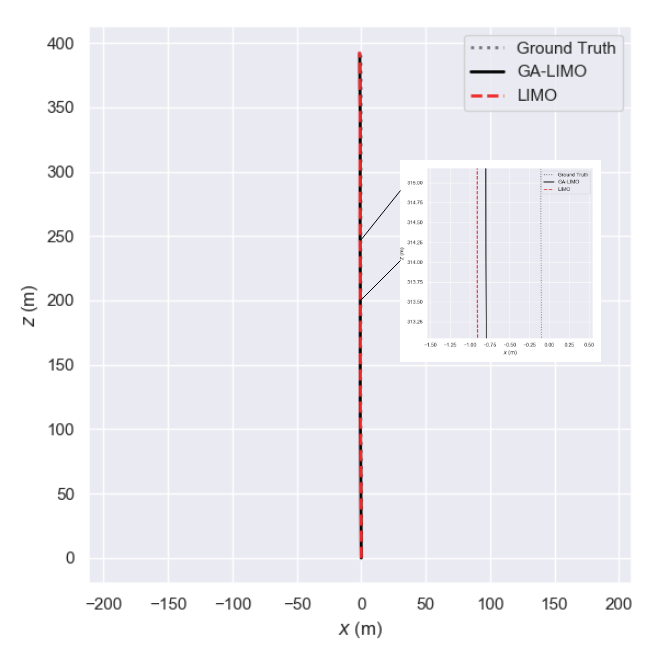}
  \label{fig:map_compare_seq_04}
  \caption{Sequence 04 trajectories showing GA-LIMO closer to ground truth.}
  \end{subfigure}
  \caption{Trajectory comparison when GA-LIMO was run as in Algorithm \ref{euclid}. In all three sequences GA-LIMO performs better than LIMO.}
  \label{fig:combined_seq_04_maps}
\end{figure}

\begin{table}[htbp]
\centering
\begin{tabular}{||c c c||} 
 \hline
 Parameters & LIMO & GA-LIMO \\ [0.5ex] 
 \hline\hline
 $\delta$ & 0.95 & 0.958\\ 
 $\epsilon_{near}$ & 400 & 999\\
 $\epsilon_{middle}$ & 400 & 593\\
 $\epsilon_{far}$ & 400 & 877\\ 
 $\mu$ & 0.9 &  0.813\\[1ex] 
 \hline
\end{tabular}
\caption{LIMO vs GA-LIMO values of parameters when GA was run on LIMO with sequence 01 individually.}
\label{table:2}
\end{table}

\begin{table}[h!]
\centering
\begin{tabular}{||c c c||} 
 \hline
 Parameters & LIMO & GA-LIMO \\ [0.5ex] 
 \hline\hline
 $\delta$ & 0.95 & 0.963\\ 
 $\epsilon_{near}$ & 400 & 999\\
 $\epsilon_{middle}$ & 400 & 554\\
 $\epsilon_{far}$ & 400 & 992\\ 
 $\mu$ & 0.9 &  0.971\\[1ex] 
 \hline
\end{tabular}
\caption{LIMO vs GA-LIMO values of parameters when GA is run on LIMO with combined sequence 01 and 04.}
\label{table:3}
\end{table}

We finally ran the system with both sequences 01 and 04 as described in Algorithm \ref{euclid}. The fitness of each evaluation is the average of translation errors of the sequences when run using the input parameters. The parameters found in GA-LIMO as shown in table \ref{table:3}, were then tested on sequences sequences 00, 01 and 04, as shown in figure \ref{fig:sequence_04_01} and \ref{fig:combined_seq_04_maps}. It is evident that GA-LIMO performed better than LIMO in all three sequences. The zoomed in figures show a closer view on one part of the trajectories. GA-LIMO trajectories are closer to the ground truth and have lesser translation errors. GA-LIMO has a translation error of 5.13\% with sequence 00, 3.59\% with sequence 01 and 0.65\% with sequence 04, in contrast with 5.77\% with sequence 00, 3.71\% with sequence 01 and 1.01\% with sequence 04 using original parameters.  

Our method helped to find common set of optimal parameters which works better and hence lead to better performance in different kinds of environments. 

\section{DISCUSSION and FUTURE WORK}
This paper shows results that demonstrated that the genetic algorithm can tune LIMO parameters to achieve better performance, reduced translation error, across a range of scenarios. We discussed existing work on VSLAM, presented an algorithm to integrate LIMO with GA to find LIMO parameters that robustly minimize translation error, and explained why a GA might be suitable for such optimization. Initial results had the assumption that GAs are a good fit for such parameter optimization, and our results show that the GA can find parameter values that lead to faster learning and better (or equal) performance. We thus provide further evidence that heuristic search as performed by genetic and other similar evolutionary computing algorithms are a viable computational tool for optimizing LIMO's performance.

\section*{APPENDIX}

Open source code for this paper is available on github: \textit{https://github.com/aralab-unr/LIMOWithGA}. The parameters used in this paper: outlier rejection quantile $\delta$; maximum number of landmarks for near bin $\epsilon_{near}$; maximum number of landmarks for middle bin $\epsilon_{middle}$; maximum number of landmarks for far bin $\epsilon_{far}$; and weight for the vegetation landmarks $\mu$, corresponds to \textit{outlier\_rejection\_quantile}, \textit{max\_number\_landmarks\_near\_bin}, \textit{max\_number\_landmarks\_middle\_bin}, \textit{max\_number\_land-} \textit{marks\_far\_bin}, \textit{shrubbery\_weight}, respectively in the code.

\bibliography{root.bib}
\bibliographystyle{IEEEtran}

\end{document}